\documentclass[submission,copyright,creativecommons]{eptcs}
\providecommand{\event}{ICLP 2023 Technical Communications} 

\usepackage{mathptmx}
\usepackage{amsmath}
\usepackage{algorithm}
\usepackage{graphicx}
\usepackage{algpseudocode}

\usepackage{listings}

\usepackage{comment}
\usepackage{cprotect}

\usepackage{fancyvrb}
\usepackage{times}
\usepackage{soul}
\usepackage{url}
\usepackage{booktabs}
\usepackage{subcaption}

\title{Reliable Natural Language Understanding with Large Language Models and Answer Set Programming}
\author{Abhiramon Rajasekharan, Yankai Zeng, Parth Padalkar, Gopal Gupta
\institute{University of Texas at Dallas\\ Richardson, USA}
\email{$\{$abhiramon.rajasekharan, yankai.zeng,
    parth.padalkar, gupta$\}$@utdallas.edu}
}
\def\titlerunning{Reliable NLU}
\def\authorrunning{Rajasekharan et al.}
\begin{document}
\maketitle

\begin{abstract}
Humans understand language by extracting information (meaning) from sentences, combining it with existing commonsense knowledge, and then performing reasoning to draw conclusions. While large language models (LLMs) such as GPT-3 and ChatGPT are able to leverage patterns in the text to solve a variety of NLP tasks, they fall short in problems that require reasoning. They also cannot reliably explain the answers generated for a given question. In order to emulate humans better, we propose STAR, a framework that combines LLMs with Answer Set Programming (ASP). We show how LLMs can be used to effectively extract knowledge---represented as predicates---from language. Goal-directed ASP is then employed to reliably reason over this knowledge. 
We apply the STAR framework to three different NLU tasks requiring reasoning: qualitative reasoning, mathematical reasoning, and goal-directed conversation. Our experiments reveal that STAR is able to bridge the gap of reasoning in NLU tasks, leading to significant performance improvements, especially for smaller LLMs, i.e., LLMs with a smaller number of parameters. NLU applications developed using the STAR framework are also explainable: along with the predicates generated, a justification in the form of a proof tree can be produced for a given output. 
 
\end{abstract}



\section{Introduction}

The long-term goal of natural language understanding (NLU)  research is to build systems that are as good as humans in understanding language. This is a challenging task since there are multiple skills that humans employ to understand a typical sentence. First, a person needs to be proficient in the language to be able to interpret the sentence and understand its surface-level meaning. Second, they need to be able to interpret the meaning of the sentence in the current context, using the commonsense knowledge they already possess. This helps resolve ambiguities in the sentence and assess if any information is missing. Third, if required, they should be able to pose a question that would seek to fill in any information that is missing. Finally, once they attain a complete understanding of the sentence, they should be able to explain what they understood. We believe that all of these skills are also important for an NLU system that seeks to reliably answer questions or hold a conversation with a human.

In recent years, Large Language Models (LLMs) have been trained on massive amounts of text extracted from the internet. They have shown language proficiency to the extent that they are able to perform reading comprehension, translate languages, and generate text to complete stories, poems, or even code (\cite{gpt3, codex}). However, they can fall short when applied to problems that require complex reasoning. When tested on commonsense reasoning or mathematics word problems, LLMs such as GPT-3 have been shown to make simple reasoning errors (\cite{gpt3-scope}). Though such errors may be mitigated with strategies such as chain-of-thought prompting (\cite{chain}), they continue to make mistakes that originate from calculation errors or missing reasoning steps in the solution, making it difficult to rely completely on such systems. While it is possible to prime LLMs to generate explanations for their answers, they sometimes generate the right explanation along with a wrong answer and vice versa (\cite{chain}). This brings into question the dependability of such explanations. The lack of a clear separation of the reasoning process also makes it difficult to assess the models' state of knowledge and identify commonsense knowledge that needs to be integrated as necessary. These shortcomings point to the need for better NLU systems that use explicit reasoning.

\begin{figure}[t]
    \centering
  \includegraphics[scale = 0.45]{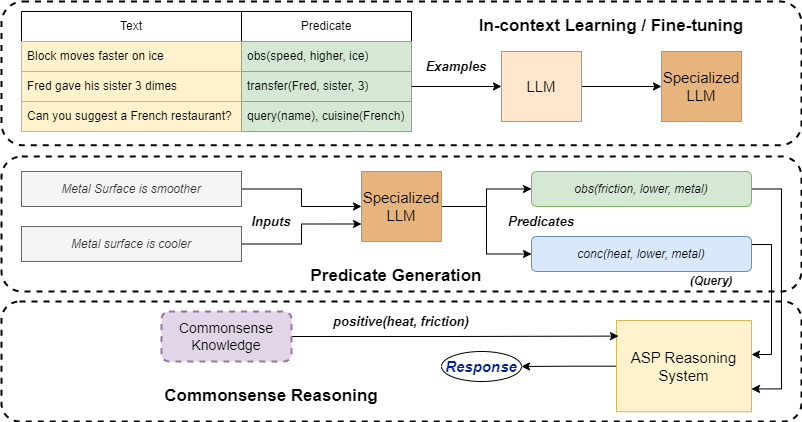}
  \caption{STAR framework Design}
  \label{fig:star}
  \vspace{-0.15in} 
\end{figure}

With this motivation, we propose the STAR (\underline{S}emantic-parsing \underline{T}ransformer and \underline{A}SP \underline{R}easoner") framework that closely aligns with the way human beings understand language. STAR maps a sentence to the semantics it represents, augments it with commonsense knowledge related to the concepts involved---just as humans do---and then uses the combined knowledge to perform the required reasoning and draw conclusions (see Figure \ref{fig:star}). The STAR framework relies on LLMs to perform semantic parsing (converting sentences to predicates that capture their semantics) and shifts the burden of reasoning to an answer set programming (ASP) system (\cite{gelfondkahl,cacm-asp}). For our experiments, we use variants of GPT-3 (\cite{gpt3}) to generate predicates from the text. LLMs can be taught to do this either using fine-tuning or in-context learning using a small number of text-predicate pairs, resulting in a `Specialized LLM'. Commonsense knowledge related to these predicates is coded in advance using ASP. Depending on the problem, a query is either pre-defined or can also be similarly generated from the problem using LLMs. The query is executed on the s(CASP) (\cite{scasp}) goal-directed ASP system against the LLM-generated predicates and ASP-coded commonsense knowledge to generate a response. 

In this paper, we use the STAR framework for three different NLU applications: (i) a system for solving qualitative reasoning problems, (ii) a system for solving math word problems, and (iii) a system representing a hotel concierge that holds a conversation with a human user who is looking for a restaurant recommendation. All three tasks require different types of reasoning. Qualitative reasoning and mathematical reasoning tasks require the system to perform a few steps of reasoning involving qualitative relationships and arithmetic operations, respectively. On the other hand, the conversation bot task requires the system to interact with the user to seek missing information, ``understand" user requirements, and reason over it. 

Our experiments involve two main variants of GPT-3; Davinci ($\sim$ 175B parameters) and Curie ($\sim$ 6.7B parameters). To measure the performance with STAR, we perform direct answer prediction using both models and compare them to the corresponding answers produced using our framework. The results show that STAR shows an increase in answer prediction accuracy and the difference is especially large for the smaller LLM, which might be weaker at reasoning. In both question-answering tasks, we are able to produce proof trees for the generated response, making them \textit{explainable}. The knowledge predicates also help us understand the shortcomings and potential design improvements, which is not possible when the models are run using the LLMs alone. In the conversation bot task that requires in-depth reasoning, we observe that STAR provides much better control wrt seeking information from the user to understand their requirements. When used on its own for the purpose of restaurant recommendation, Davinci sometimes alters restaurant information based on user interaction. However, our approach always gives restaurant suggestions faithfully based on the database of restaurants available, making it more \textit{reliable}. Since reasoning is performed using s(CASP) in our approach, we can also handle an arbitrarily large database of restaurants. This is not possible when LLMs are used end-to-end for this conversation bot, as there is a limit on the maximum prompt size. Thus, our approach can scale easily to larger restaurant databases.

\section{Background}
\label{sec:background}

\smallskip\noindent\textbf{Large Language Models:}
Until recently, transformer-based deep learning models have been applied to NLP tasks by training and fine-tuning them on task-specific datasets (\cite{pre_trained_transformers}). With the advent of Large Language Models, the paradigm changed to teaching a language model any arbitrary task using just a few demonstrations, called \textit{in-context learning}. Brown et al. (\cite{gpt3}) introduced an LLM called GPT-3 containing approximately 175 billion parameters that have been trained using a massive corpus of filtered online text, on which the well-known ChatGPT is based (\cite{chatgpt})). 
The model was able to perform competitively on several tasks such as question-answering, semantic parsing (\cite{LLM_sem_parse}), and machine translation. However, such LLMs tend to make simple mistakes in tasks such as semantic (commonsense) and mathematical reasoning (\cite{gpt3-scope, chain}). 

In our work, we use GPT-3 for semantic parsing and leave the reasoning part to ASP. We theorize that given the vast pre-training they go through, LLMs can be used to automatically extract knowledge inherent in the text, just like humans do. 
Our experiments confirm that Davinci and Curie are able to extract such knowledge as predicates from sentences---with high accuracy---after learning from a few example demonstrations. 
Thus, our experiments show that LLMs are able to extract, what linguists call, the \textit{deep structure} of a sentence, given a sentence's \textit{surface structure}. 

\medskip\noindent\textbf{Answer Set Programming and the s(CASP) system:}
The s(CASP) system (developed by Arias et al.\cite{scasp}) is an answer set programming (\cite{cacm-asp}) system that supports predicates, constraints over non-ground variables, uninterpreted functions, and, most importantly, a top-down, query-driven execution strategy.
These features make it possible to return answers with non-ground variables (possibly including constraints among them) and compute partial models by returning only the fragment of a stable model that is necessary to support the answer to a given query.
The s(CASP) system supports constructive negation based on a disequality constraint solver and unlike Prolog's negation as failure and ASP's
default negation, %
{\tt not p(X)} can return bindings for \texttt{X} on success, i.e., bindings for which the call \texttt{p(X)} would have failed. Additionally, s(CASP) system's interface with a constraint solver (over reals) allows for sound non-monotonic reasoning with constraints (useful for solving algebra problems in one of the NLU applications we discuss later).

Complex commonsense knowledge can be represented in ASP and the s(CASP) query-driven predicate ASP system can be used for querying it (\cite{gupta-csr,murder-trial,gelfondkahl}). Commonsense knowledge can be emulated using (i) default rules, (ii) integrity constraints, and (iii) multiple possible worlds~(\cite{gelfondkahl,gupta-csr}). Default rules are used for jumping to a conclusion in the absence of exceptions, e.g., a bird normally flies unless it's a penguin. 
Default rules with such exceptions represent an elaboration-tolerant way of representing knowledge~(\cite{gelfondkahl}). 

{\tt
flies(X) :- bird(X), not abnormal\_bird(X).

abnormal\_bird(X) :- penguin(X).} 

\smallskip 

\noindent Integrity constraints allow us to express impossible situations and invariants. For example, a person cannot sit and stand at the same time.

{\tt false :- person(X), sit(X), stand(X).}

\smallskip

\noindent Finally, multiple possible worlds allow us to construct alternative universes that may have some of the parts common but other parts inconsistent. For example, the cartoon world of children's books has a lot in common with the real world (e.g., birds can fly in both worlds), yet in the former birds can talk like humans but in the latter they cannot. 

Default rules are used to model a bulk of our commonsense knowledge. Integrity constraints help in checking the consistency of the information extracted. Multiple possible worlds allow us to perform assumption-based reasoning (for example, knowing that ``Alice loves Bob", we could assume that either Bob also loves Alice or he does not). 

A large number of commonsense reasoning applications have already been developed using ASP and the s(CASP) system (\cite{blawx, logical-english, chef, murder-trial}). In the three applications reported in this paper, we have kept the  commonsense reasoning component simple, as our main goal is to illustrate our framework for combining LLMs and ASP to develop NLU applications that are explainable and reliable. Because of the use of ASP, it is also possible to detect inconsistencies or biases in the text by reasoning over the predicates extracted. Justification for each response can also be given, as the s(CASP) system can generate justifications as proof trees (\cite{scasp-justification}).

\section{Qualitative Reasoning}
\label{sec:qualitative}
Qualitative reasoning tests a model's ability to reason about the properties of objects and events in the World. Tafjord et al. (\cite{quarel}) introduced the QuaRel dataset in order to test question answering about qualitative relationships of a set of physical properties, which forms a perfect test-bed for our approach. 
Our experimental results show that the STAR framework significantly improves the model accuracy compared to the cases where the LLMs are applied directly to question answering.

\vspace{-0.12in} 
\subsection{The QuaRel Dataset} 
The QuaRel dataset consists of 2771 questions designed around 19 different properties such as `friction', `heat', `speed', `time', etc. In order to answer these questions, one must account for the correlation between these properties. Each question has a certain observation made about the two worlds where a property has a higher (or lower) value in one world compared to the other. Based on this observation, a (commonsense) inference needs to be drawn about other related properties described in the two worlds. This inference helps pick one of the two choices as the answer for the given question \cite{quarel}. 

A question from the dataset is given in example \ref{verb:quarel-example}. In this example, the two worlds are 'Carpet' and 'Floor'. The observation made is that the \textit{distance} traveled by a toy car is more in world1 (floor). From this, the model needs to infer that the resistance or \textit{friction} would be higher in world2 (carpet), which should lead to picking option A as the answer.
\medskip 

\noindent 
\textbf{Example 3.1:}
{\small
\begin{verbatim}
Question: Alan noticed that his toy car rolls further on a wood 
floor than on a thick carpet. This suggests that:
(world1: wood floor, world2: thick carpet)
       (A) The carpet has more resistance (Solution)
       (B) The floor has more resistance
\end{verbatim}\label{verb:quarel-example}
}
\noindent Along with each question, Tafjord et al. provide a logical form that captures the semantics of the question and we use it to extract the predicates needed for our method (\cite{quarel}). For the above question (example \ref{verb:quarel-example}), the logical form given is as follows:

\vspace{-0.15in}
{\small
\begin{equation}\label{eqn:quarel-example}
    qrel(distance, higher, world1) \rightarrow qrel(friction, higher, world2)\ ;\ qrel(friction, higher, world1)
\end{equation}
}

\noindent The predicate $qrel(distance,higher,world1)$ refers to the observation that the \textit{distance} is higher in world1, while $qrel(friction,higher,world2)$ and $qrel(friction,higher,world1)$ refer to the conclusions drawn in the two answer options, respectively. 

\vspace{-0.1in} 
\subsection{Predicate Generation Step}
We use GPT-3 to convert the Quarel dataset's natural language question (including the two answers) into appropriate predicates. We \textit{fine-tune} the two GPT-3 model variants named Davinci and Curie (\cite{gpt3}) on the QuaRel dataset, instead of just using \textit{in-context learning}\footnote{Fine-tuning an LLM involves using additional training data to refine the LLM for the task at hand; in-context learning refers to giving some examples from the training data, along with the question posed, to the LLM as a part of its input.}. Fine-tuning performs better since the models can learn from all the examples in the training set. Such data is available only for this task.
Our input prompt consists of the question (including answer options), followed by the world descriptions. The world descriptions are included to enable the model to link the two worlds to the ones in the predicates (\textit{obs} and \textit{conc}) that are generated in the output. The prompt and completion formats for fine-tuning are given below:

\noindent \textbf{Prompt format:}

{\small {\small\tt <Question-Answers>$\backslash$n\ world1:<world1>$\backslash$n world2:<world2>$\backslash$n$\backslash$n\#\#$\backslash$n$\backslash$n}}

\noindent \textbf{Completion format:}

\centerline{{\small{\small\tt obs(<p>, <h/l>, <w1/w2>)} $\rightarrow${\tt \ conc(<p>, <h/l>, <w1/w2>)\ ;}}}

\centerline{\small{{\small\tt conc(<p>, <h/l>, <w1/w2>)\ <EOS>}}}

\noindent where p is the \textit{property} involved, h/l is the relation which can be either \textit{higher} or \textit{lower} and w1/w2 is either \textit{world1} or \textit{world2}. After fine-tuning on the training set using the prompt and completion pairs, we use the prompt to generate the completion during testing. The {\tt <EOS>} token helps cut off the generation when apt, avoiding completions that are either too long or too short. The extracted \textit{obs} and \textit{conc} predicates are then used by the logic program to determine the correct answer.

\vspace{-0.1in} 
\subsection{Commonsense Reasoning Step}
The commonsense knowledge required to answer the questions is encoded in ASP as facts and rules. First, we ground the 19 properties using facts such as,
{\small\begin{verbatim}
property(friction).     property(heat).     property(speed).
\end{verbatim}}
\noindent Next, we define the relationships between the properties, including their positive correlations (denoted as qplus), negative correlations (denoted as qminus) and symmetry,
{\small\begin{verbatim}
qplus(friction, heat).            qminus(friction, speed).    
qplus(speed, distance).           qminus(distance, loudness).
positive(X, Y) :- qplus(X, Y).    negative(X, Y) :- qminus(X, Y).
positive(X, Y) :- qplus(Y, X).    negative(X, Y) :- qminus(Y, X).
\end{verbatim}}

In the QuaRel dataset, we are only dealing with two worlds. Hence, if a property P is higher in world1, it must be lower in world2 and vice versa. We capture this logic using the \textit{opposite} predicates and the rules below:

\vspace{-0.05in}
{\small\begin{verbatim}
    opposite_w(world1,world2).       opposite_v(higher,lower).
    opposite_w(world2,world1).       opposite_v(lower,higher).
    conc(P, V, W) :- obs(P, Vr, Wr), property(P), 
             opposite_w(W,Wr), opposite_v(V,Vr).
\end{verbatim}}
\vspace{-0.05in} 

\noindent In order to capture the relationship between each pair of properties, we need to account for 4 different cases that may arise. If properties P and Pr are positively correlated, then (i) if  P is higher in world W, Pr must also be higher in W, and (ii) if P is higher in world W, Pr must be lower in the other world Wr. Similarly, if P and Pr are negatively correlated, then (i) if  P is higher in world W, Pr must be lower in W, and (ii) if P is higher in world W, Pr must be higher in the other world Wr. Note that the higher/lower relations may be swapped in all cases above.
These 4 possible scenarios can be encoded in logic using the following rules:

\vspace{-0.05in}
{\small\begin{verbatim}
    conc(P,V,W) :- obs(Pr,V,W), property(P), property(Pr), 
            positive(P,Pr).          
    conc(P,V,W) :- obs(Pr,Vr,Wr), property(P), property(Pr), 
            opposite_w(W,Wr), opposite_v(V,Vr), positive(P,Pr).           
    conc(P,V,W) :- obs(Pr,Vr,W), property(P), property(Pr), 
            opposite_v(V,Vr), negative(P,Pr).          
    conc(P,V,W) :- obs(Pr,V,Wr), property(P), property(Pr), 
            opposite_w(W,Wr), negative(P,Pr).
\end{verbatim}}

\vspace{-0.05in}

\noindent Using this knowledge base, asserting a fact as an observation (\textit{obs}) allows us to check for the correct conclusion (\textit{conc}) that is entailed. For the example question in example ~\ref{verb:quarel-example}, we can arrive at the answer by checking for entailment of the two possible conclusions as shown:

$assert(obs(distance,\ higher,\ world1)),\ conc(friction,\ higher,\ world2). \rightarrow\ True$

\noindent and 

$assert(obs(distance,\ higher,\ world1)),\ conc(friction,\ higher,\ world1). \rightarrow\ False$




\subsection{Results and Evaluation}
We compare the results of our models to those reported by Tafjord et al. (\cite{quarel}) in Table \ref{table:quarel-results}. Accuracy for four QuaRel datasets is considered ($QuaRel^F$ refers to the subset of the dataset which only focuses on friction-related questions). The first 8 rows show the accuracy of the baseline models proposed in the QuaRel paper. Curie-Direct and Davinci-Direct rows report the performance of Curie and Davinci models which directly predict the answer after fine-tuning on the QuaRel's training set. The Curie-STAR and Davinci-STAR rows show the performance for our approach, i.e., first generating the predicates and then reasoning using ASP and commonsense knowledge. The values in bold represent the highest accuracy values obtained for each dataset.

The results show a large improvement in the accuracy of the Curie model on all four QuaRel datasets. Table 1 shows that Davinci-STAR either matches or exceeds the performance of Davinci-Direct on three of the four QuaRel datasets. However, interestingly, we see that Davinci-Direct outperforms Davinci-STAR on the QuaRel-Dev dataset. Since our framework is explainable, we were able to analyze the cases where our approach makes a mistake. We found that the LLM sometimes generates properties that are not in the domain for some predicates (such as 'smoke' instead of 'heat' since the question mentions smoke). We hypothesize that this is because similar examples were not seen during training. QuaRel has a larger number of questions based on friction, which we believe led to our framework performing better on $QuaRel^F$ datasets (which contain solely friction-based questions). Similarly, adding more examples for other properties might help our framework bridge the gap for other properties.
Clearly, there is a stark difference between Curie and Davinci when used with our framework. We infer from this that while Davinci has some ability to reason, Curie lacks the reasoning skill required for the task and our approach helps bridge this reasoning gap.

{\footnotesize 
\begin{center}
\begin{table}[H]
\caption{Comparison of accuracy of models on the QuaRel Dataset (Qualitative Reasoning)}
 \begin{minipage}{\textwidth}
 \begin{center}
    \begin{tabular}{llrrrr}
\hline\hline
No.&Model&
    QuaRel Dev&
    QuaRel Test&
    $QuaRel^F$ Dev&
    $QuaRel^F$ Test\\[0.5ex]
\midrule
1.&Random & 50.0 & 50.0 & 50.0 & 50.0 \\
2.&Human & 96.4 & - & 95.0 & - \\
3.&IR & 50.7 & 48.6 & 50.7 & 48.9 \\
 4.&PMI & 49.3 & 50.5 & 50.7 & 52.5 \\
  5.&Rule-Based & - & - & 55.0 & 57.7 \\
  6.&BiLSTM & 55.8 & 53.1 & 59.3 & 54.3 \\
  7.&QUASP & 62.1 & 56.1 & 69.2 & 61.7 \\
  8.&QUASP+ & 68.9 & 68.7 & 79.6 & 74.5 \\
  9.&Curie-Direct & 67.6 & 63.5 & 45.7 & 52.7 \\
 10.&Curie-STAR (ours) & 86.2 & 85.2 & 87.9 & 85.9 \\ 
 11.&Davinci-Direct & \textbf{93.1} & \textbf{90.5} & 90.0 & 91.3 \\
 12.&Davinci-STAR (ours) & 90.6 & \textbf{90.5} & \textbf{90.6} & \textbf{93.5}\\
      \hline\hline
    \end{tabular}
    \end{center}
  \end{minipage}
  \label{table:quarel-results}
\end{table}
\end{center}
}


\section{Solving Word Problems in Algebra}
\label{sec:math}
Solving word problems in algebra requires extracting information from the question (interpreting its language) and performing mathematical reasoning to come up with an answer. Hence, it forms a good experiment to test our framework. We choose a specific type of addition and subtraction problems from the dataset used by Koncel-Kedziorski et al. (\cite{MWP}).
We define the predicates {\tt has/4}, {\tt transfer/5} and {\tt total/4} as shown below to encode the knowledge in the problems:

{\small\tt has(entity, quantity, time\_stamp, k/q).}
    
{\small\tt transfer(entity1, entity2, quantity, time\_stamp, k/q).}
   
{\small\tt total(entity, quantity, time\_stamp, k/q).}

\noindent The predicate {\tt has/4} defines that an {\tt entity} has a certain {\tt quantity} of some objects, at a particular {\tt time\_stamp}. The {\tt transfer/5} predicate defines that an {\tt entity1} has transferred a certain {\tt quantity} of objects, to {\tt entity2} at a particular {\tt time\_stamp}. Finally, the {\tt total/4} predicate defines that an {\tt entity} has a total amount of some objects equal to the {\tt quantity}, at a particular {\tt time\_stamp}. The last term in each predicate is either the current knowledge (denoted as {\tt k}) or a placeholder for the query (denoted {\tt q}). We design these based on what information a human might glean from the problem in order to solve it.

The computation of the answer is done by simple s(CASP) rules. The rules are not shown due to lack of space and can be thought of as commonsense knowledge required to solve simple Algebra word problems given the {\tt has/4}, {\tt transfer/5}, and {\tt total/4} predicates.
An example problem and corresponding predicates generated to represent the knowledge are shown below.

\medskip\noindent\textbf{Ex 1:}
\textit{Joan found 70 seashells on the beach. Joan gave Sam some of her seashells. Joan has 27 seashells left. How many seashells did Joan give to Sam?}

{\tt has(joan,70,0,k). transfer(joan,sam,X,1,q). has(joan,27,2,k). }





\medskip \noindent 
Following the STAR approach, we convert the knowledge in the chosen algebraic problem to the predicates defined above using an LLM. The predicates thus obtained (including the query) along with the rules then constitute the logic program. The query predicate is then executed against the program to solve the word problem.

\subsection{Experiments and Results}
Our dataset contains $91$ problems drawn from a collection of word problems provided by Koncel-Kedziorski et al. (\cite{MWP}). Since we hand-craft the rules for the domain, we select a set of problems that have a similar logic that we can encode. We use text-davinci-003 which is the most capable GPT-3 model for in-context learning. We did not use the text-curie-001 model as done in the qualitative reasoning experiment because the model requires fine-tuning on a larger set of questions to be effective. We provide a context containing a few problems with their corresponding predicates to the GPT-3 models and then use each problem as a prompt along with the context for the model to generate the facts and the query predicate(s) corresponding to the new problem. We then use the commonsense rules we defined along with the generated predicates (facts) as the logic program and query the program using the query predicate. We then compare the answer generated by the logic program with the actual, human-computed answer for each problem. As a baseline, we use the GPT-3 model for direct answer prediction. Here, with the 
algebra problems as the context, we provide the correct answer as the expected completion.

For our experiments, we initially started with a smaller context of $12$ problems and examined the mistakes the LLM was making in generating the predicates. Since our approach is explainable (unlike the direct answer prediction approach), we were able to analyze the mistakes and added more problems to the context that might fix them. Repeating this process a few times, we end up with $24$ problems as the final context for the GPT-3 model. Results of our experiments are shown in table \ref{table:math_results}.

{\footnotesize 
\begin{center}
\begin{table}[H]
\caption{Performance comparison between the baseline model and our approach}
 \begin{minipage}{\textwidth}
 \begin{center}
    \begin{tabular}{lr}
\hline\hline
Model&
Accuracy\\
\midrule
text-davinci-003-Direct & $\textbf{1.00}$ \\
text-davinci-003-STAR & $\textbf{1.00}$\\
      \hline\hline
    \end{tabular}
    \end{center}
  \end{minipage}
\label{table:math_results}
\end{table}
\end{center}
}

Both text-davinci-003-Direct and text-davinci-003-STAR result in $100\%$ accuracy on the test set of $67$ problems. Our experiments show that algebraic word problems that require simple reasoning can be solved easily by large LLMs. While our STAR approach achieves the same accuracy, it can also generate justification, making it explainable. Below we show a justification tree generated by s(CASP) (\cite{scasp-justification}) for the problem shown in Example 1 above, where the computed answer is 43.

{\small
\begin{verbatim}
JUSTIFICATION_TREE:
transfer(joan,sam,43,1,q) :-
    has(joan,70,0,k),
    has(sam,27,2,k),
    43 #= 70-27.
global_constraint. 
\end{verbatim}
}
A major weakness of LLMs is that they perform poorly on mathematical tasks. Since the computation in our approach is done by the reasoning process externally (as seen above), it circumvents any potential calculation mistakes the LLMs might make. The set of problems we selected ended up being simple enough for the baseline model to also learn. We believe that the performance difference will be evident with more complex algebraic problems, similar to our experiments on qualitative reasoning. We plan to explore this in our future work.

\section{Concierge Bot: A Goal-Directed Conversational Bot}
\label{sec:conversation}
In this section, we describe the conversational bot that we have developed using our STAR framework. We first leverage LLMs to extract implicit and explicit knowledge predicates from a user's dialogue and then reason over the extracted predicates using s(CASP) to satisfy the user's request.
To illustrate the applicability of our method to conversational AI, we narrow the domain of the bot to a very specific problem: restaurant recommendations. This ensures that commonsense knowledge that needs to be modeled is within a limited domain that can be expanded later as necessary. We call this bot the concierge bot.
Since the conversational bot is an interactive application, the commonsense reasoning involved is more complex than the other two tasks. Here, the reasoner also has to keep track of the state of the conversation. In order to make a restaurant recommendation, the bot needs to ask the user for  information (preferred cuisine, price range, etc.) and recommend a restaurant once all information is procured. At any stage, the bot can explain why it's asking for a given piece of information as well as the reason for its recommendation.
The benefits of the STAR framework are more apparent in this application. The bot can ask user questions to acquire more information when missing or to verify if any errors are made in predicate generation. We show an example transcript of an interaction session below of the concierge bot with a human user. The English responses are generated using GPT-3 again after learning from predicate-text demonstrations (essentially, using the reverse of the 
text-to-predicate generation process). 

{\small
\begin{verbatim}
Bot:   Hi, what can I assist you with?
Human: Can you help me find a place for food with curry? 
       I don't want a pricey one.
Bot:   Are you looking for a place with a specific customer rating?
Human: No, I don't mind the rating.
Bot:   What kind of restaurant would you like to visit?
Human: A normal restaurant.
Bot:   Would you like somewhere family-friendly for tonight?
Human: No. Just for myself.
Bot:   Unfortunately, we cannot provide the results to your request.
Human: How about one with a high price? 
       But it should be then at least above average quality.
Bot:   Are you looking for a Thai restaurant? 
       The Rice Boat may be suitable but it can be pricey and 
       customer feedback has been mixed. Additionally, it may 
       not have the best atmosphere for families.
Human: Yes, that's what I need! Tell me where it is.
Bot:   The Rice Boat, located on 901 Bangkok Road, has an average 
       customer rating and offers Thai cuisine at a premium cost. 
       Unfortunately, it is not suitable for children.
Human: Great! Thank you for the service!
Bot:   It's no problem, I'm happy to assist.
\end{verbatim}
}

\subsection{Predicate Generation using LLMs}
\label{sec:convbot_test}

We adapt examples from the E2E dataset released by Novikova et al. (\cite{novikova2017e2e}) for our use case. We employ in-context learning where we give the LLM examples of pairs containing a natural language sentence and the corresponding restaurant predicates. 
%
We use the following six predicates from the dataset: \textbf{restaurant-name}, \textbf{eatType} (as establishment type), \textbf{food} (as cuisine), \textbf{priceRange}, \textbf{customerRating}, \textbf{familyFriendly}.

We use the STAR framework with in-context learning where we provide the GPT-3 model with 11 selected examples from the dataset, which covers all the predicates along with their possible arguments. This ensures that the LLM is aware of every possible predicate as well as every possible argument value these predicates can take. To assess the viability of LLMs for the predicate generation task, we tested the model using the first 500 examples in the E2E training set and obtained an accuracy of 89.33\%. The accuracy metric we use is designed to account for the generation of correct predicates as well as arguments. The high predicate generation accuracy supports the feasibility of using our STAR framework for the concierge bot. Our framework can similarly be applied, to build any robust domain-specific conversational bots such as a front desk office receptionist or an airline reservation assistant.

\subsection{Concierge Bot System Construction}
To make GPT-3 better understand the meaning of each predicate, we first change the predicate names in E2E as follows: \textbf{restaurnt-name}, \textbf{typeToEat}, \textbf{cuisine}, \textbf{priceRange}, \textbf{customerRating}, \textbf{familyFriendly}. We also add two predicates \textbf{address} and \textbf{phoneNumber} to record the location and contact information for the user's query. An external predicate \textbf{prefer} is also added to capture the user's preference (such as curry, spicy, etc.) The information asked by the user is expressed by the value ``query". We specialized GPT-3 with about a dozen example sentences along with the corresponding predicate(s). Below we show some examples of the sentences and the predicates generated after this specialization. 

{\small 
\begin{verbatim}
Sentence: Fitzbillies coffee shop provides a kid-friendly venue for 
    Chinese food at an average price point in the riverside area.
    It is highly rated by customers.
Predicates: restaurant-name(Fitzbillies), typeToEat(coffee shop),
            cuisine(Chinese), priceRange(moderate),
            customerRating(high), familyFriendly(yes)

Sentence: Can you find a place for food at a low price? Both English
    and French cuisine is fine for me.
Predicates: restaurant-name(query), cuisine([Engish, French]), 
            priceRange(cheap)
\end{verbatim}


} 
\begin{figure}
    \centering
    \includegraphics[width=0.8\linewidth]{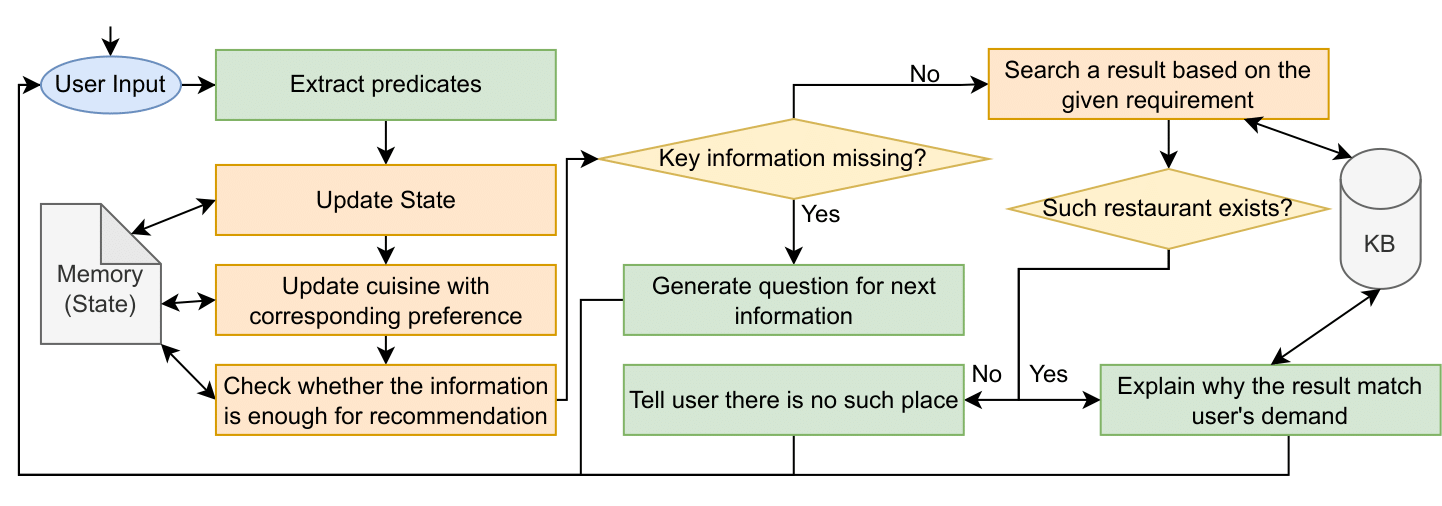}
    \caption{The framework of the reasoning system in Concierge Bot. The green boxes indicate the steps done by LLMs and the orange ones indicate the steps done by s(CASP).}
    \label{fig:chatbot_flow}
\end{figure}
Commonsense knowledge involved in making a restaurant recommendation is coded using s(CASP). The interactive bot will take in the user's response and convert it to predicates using GPT-3. The predicates become part of the state. At this stage, we check for user preference. For example, if the user wants curry, Indian and Thai cuisine would be automatically added to the state through appropriate rules. The bot then examines the state to assess if all the information needed is present so that it can make a recommendation and if not, it will generate a question to ask the user for that information. This logic, shown in Figure \ref{fig:chatbot_flow}, can be thought of as a state machine and has been referred to as a conversational knowledge template (CKT) by Basu et al. (\cite{ckt}). The concierge bot determines which predicates are missing in its state to make a recommendation. One of the missing predicates is then selected and a query is created using it. Note that we use GPT-3 again to generate natural-sounding text from the predicate(s) corresponding to the response that our bot computes. The users can also change their preferences during the conversation and our bot can handle that.
Take the conversation mentioned above as an example. When the user asks ``Can you help me find a place for food with curry? I don't want a pricey one.", following predicates are generated by the GPT-3 text-davinci-003 model: \textit{``restaurant-name(query), prefer(curry), priceRange([cheap, moderate])"}. The predicates are then added to the memory of the bot, where the log of the user requirements of the current conversation is stored. Note that the predicates \textit{``prefer(curry)"} and \textit{ ``cuisine([indian,thai])"} are also added to the state using the specific commonsense rule we wrote for this purpose (not shown due to lack of space).
%
%
After the state gets updated, the bot checks to see whether any other key information is missing using another commonsense rule (not shown also due to lack of space).
%
%
If no information is missing, the bot comes to the recommendation step where it searches for the appropriate restaurant in its database, consistent with the user's input, and outputs it as a recommendation. 

Since we use GPT-3 to generate predicates from user utterances, the user has complete freedom in terms of what they say. GPT-3 is able to extract the appropriate predicates from the corresponding deep form of the sentence. 
GPT-3 is also used to generate the natural language response of the bot. As a result, the interaction appears quite natural, as the example above shows.

\subsection{Experiments}
To explore GPT-3's ability to generate goal-directed conversation without the reasoning system that STAR contains, we tested the conversation directly on GPT-3 with the sample conversations as input. We showed GPT-3 three full paragraphs of conversations and asked GPT-3 to generate a similar conversation according to the user input. The database of restaurants used by our concierge bot is also included in the input to the GPT-3 model. 
An example interactive conversation directly with GPT-3 is as follows:

{\small
\begin{verbatim}
Bot:   Hello, what can I do for you today?
Human: I'd like to have a family meal with my children.
Bot:   What type of cuisine would you like?
Human: Mexican food is suitable for me.
Bot:   What is your desired price range for the place?
Human: At a higher price, please.
Bot:   I recommend The Wrestlers, a coffee shop serving Mexican 
       cuisine at a moderate price range. It is family-friendly
       and has an average customer rating. The address is 470 Main Rd.
\end{verbatim}
}

The responses given by GPT-3 in the above conversation are correct except for the price range. In the given database, the restaurant recommended only serves cheap food. Hence, GPT-3 modified the information to align with the user's request. GPT-3 also follows the given examples and asks about the cuisine and price, but does not request other information like our STAR framework does. This is because these questions are not motivated by missing information, unlike in our approach. This example shows that although GPT-3 used on its own as a conversational bot is able to generate natural-sounding sentences fluently, it is unreliable and does not understand the knowledge given. Bots developed using our framework do not face such problems because they employ explicit commonsense reasoning. The methodology we use to build the concierge bot is explained in more detail in the paper Zeng et. al. ~\cite{auto-concierge}.

\section{Related Work}
\label{sec:related}
A recent line of research on improving the reasoning capabilities of LLMs focus on prompt engineering. Wei et al. (\cite{chain}) show that generating a chain of thought before the answer leads to a significant improvement in performance in a variety of reasoning tasks. However, in some cases, a wrong reasoning chain can lead to the right answer or vice versa. Zelikman et. al. (\cite{star}) extend this by generating rationales using a self-taught approach. While the above approaches focus primarily on machine learning, our approach instead relies on s(CASP) to perform reasoning explicitly. This explicit reasoning is not only more reliable but is also explainable. Our approach falls into the line of Neuro-symbolic research that does heavy reasoning and light learning as categorized in the survey by Hamilton et al. (\cite{nesy-survey}). Typically such works try to integrate neural and symbolic components (\cite{neurasp, logicnn}). In contrast, we separate both components by using LLMs for predicate extraction and the s(CASP) system for reasoning.

This paper follows our earlier research where we advocate a combination of machine learning and commonsense reasoning to carry out intelligent tasks in a human-like manner (\cite{aqua, square, auto-discern}). In the SQuARE question-answering system (\cite{square}), knowledge was extracted using the Stanford CoreNLP Parser (\cite{corenlp}) and then mapped to templates from VerbNet. This method was only usable for simple sentences such as for the bAbI dataset. Along similar lines, Tafjord et al. (\cite{quarel}) report models that convert problems in English to logical forms, which are then processed using a custom interpreter. The semantic parsers discussed in their paper are variations of LSTMs that 
generate CFG-like grammar rules which create the logic form. The main bottleneck of these two approaches was in the parsers used. In our work, we use LLMs that can extract predicates from arbitrary sentences which makes our approach applicable to more complex problems and translates to better performance. The closest works to ours are along the lines of Chen et al. (\cite{prog-prompt}) and Gao et al. (\cite{pal}) who use LLMs to generate program steps which are then executed in a programming language such as Python. Here the program steps still need to be generated by LLMs completely. Instead, our work delegates the entire reasoning task to the s(CASP) system. Since s(CASP) is designed for complex reasoning, our research can be extended more easily to complex text-based reasoning problems. Similar to our conversation bot, Inclezan et al. (\cite{restKB}) use an ASP-based action language to reason in a restaurant setting, but it is not an interactive system like our work.

\section{Conclusions and Future Work}
\label{sec:conc}
In this paper, we described the STAR framework that combines LLMs and ASP for NLU tasks. We show that our system is reliable and explainable using three different reasoning tasks. For the qualitative reasoning task, STAR  outperforms purely LLM-based approaches and advances the state-of-the-art wrt performance on most datasets in QuaRel. The performance difference is more significant for Curie, indicating that it helps bridge the reasoning gap in smaller LLMs. In all three tasks, STAR can explain its reasoning process by producing a justification tree. In the LLM-only approach for developing a concierge bot, we noticed that the LLM mixes up information collected during the conversation and leads to incorrect suggestions, while our STAR-based approach stays faithful to the information given in the restaurant database.  Our approach also allows for holding long, interactive, and meaningful conversations.

The potential applications of our STAR framework are very broad. It can help in any NLU application that requires reasoning about knowledge in text or utterances. Some examples are, automatically extracting formal software requirements from textual specifications, building conversational agents for other domains, and reliable machine translation.
We believe that performance improvement using STAR will be more pronounced for problems that require complex reasoning. We also plan to develop a general commonsense knowledge base that applications developed using the framework can employ. 

\smallskip 
\noindent{\bf Acknowledgment:}
Authors acknowledge support from NSF grants IIS 1910131, IIP 1916206, and a DARPA ARCOS award. 
We are grateful to Farhad Shakerin, Kinjal Basu, Joaquin Arias, Elmer Salazar, and members of the UT Dallas ALPS Lab for their discussions.

\vspace{-0.15in} 

\bibliographystyle{eptcs}
\bibliography{generic}
\end{document}